\documentclass[conference]{IEEEtran}

\usepackage{graphicx}
\graphicspath{ {./Figures/} }
\IEEEoverridecommandlockouts
\usepackage[american]{babel}
\usepackage[utf8]{inputenc}
\usepackage[T1]{fontenc}
\usepackage{newfloat}
\usepackage{booktabs}
\usepackage{amssymb, amsmath, array, bm, algorithm, standalone,csquotes,textcomp, amssymb,amsfonts, tabularx}
\usepackage{gensymb, wasysym} 
\usepackage{graphicx, xcolor, subcaption, soul}
\usepackage{siunitx, nameref, zref-xr}
\usepackage{algorithmic}
\usepackage{comment}
\usepackage[]{mdframed}
\zxrsetup{toltxlabel}
\usepackage{microtype}
\usepackage[nolist]{acronym}
\definecolor{myBlue}{HTML}{03456A} 
\definecolor{Gray}{RGB}{230,230,230}
\usepackage{hyperref}
\hypersetup{
    colorlinks=true,
    linkcolor=myBlue,
    citecolor=myBlue,
    filecolor=myBlue,      
    urlcolor=myBlue,
    pdftitle={main},
    pdfpagemode=FullScreen,
    }
\usepackage[nameinlink]{cleveref}

\newcommand\mailto[1]{\href{mailto:#1}{#1}}

\let\oldcite\cite
\renewcommand*\cite[1]{\,\oldcite{#1}}
\def\BibTeX{{\rm B\kern-.05em{\sc i\kern-.025em b}\kern-.08em
    T\kern-.1667em\lower.7ex\hbox{E}\kern-.125emX}}

\begin{document}
\begin{acronym}[MPC]
\acro{DMS}{Distributed Manipulation System}
\acro{DOF}{Degrees of Freedom}
\acro{NCA}{Neural Cellular Automata}
\acro{CA}{Cellular Automata}
\acro{CoM}{Center of Mass}
\acro{GC}{Geometric Center}
\acro{IMU}{Inertial Measurement Units}
\acro{NN}{Neural Network}
\acro{CV}{Computer Vision}
\acro{PCB}{Printed Circuit Board}
\acro{PET}{Polyethylene Terephthalate}
\acro{MSE}{Mean Square Error}
\acro{CNN}{Convolutional Neural Network}
\end{acronym}

\title{Neural Cellular Automata for Decentralized Sensing using a Soft Inductive Sensor Array for Distributed Manipulator Systems\\
}
\author{
Bailey Dacre,
Nicolas Bessone,
Matteo~Lo Preti,
Diana~Cafiso,
Rodrigo Moreno,\\
Andrés Faíña,
and Lucia~Beccai
\thanks{M. Lo Preti, D. Cafiso, and L. Beccai are with the Istituto Italiano di Tecnologia, Genova, 16163, IT (e-mail:  \mailto{matteo.lopreti@iit.it},\mailto{diana.cafiso@iit.it}, \mailto{lucia.beccai@iit.it}).}
\thanks{B. Dacre, N. Bessone, R. Moreno, and A. Faíña are with the IT University of Copenhagen (ITU), København, DK (e-mail: \mailto{baid@itu.dk}, \mailto{nbes@itu.dk}, \mailto{rodr@itu.dk}, \mailto{anfv@itu.dk})}
\thanks{This work has been submitted to the IEEE for possible publication. Copyright may be transferred without notice, after which this version may no longer be accessible.}
}
\maketitle

\begin{abstract}
In Distributed Manipulator Systems (DMS), decentralization is a highly desirable property as it promotes robustness and facilitates scalability by distributing computational burden and eliminating singular points of failure. However, current DMS typically utilize a centralized approach to sensing, such as single-camera computer vision systems. This centralization poses a risk to system reliability and offers a significant limiting factor to system size. In this work, we introduce a decentralized approach for sensing and in a Distributed Manipulator Systems using Neural Cellular Automata (NCA). Demonstrating a decentralized sensing in a hardware implementation, we present a novel inductive sensor board designed for distributed sensing and evaluate its ability to estimate global object properties, such as the geometric center, through local interactions and computations. Experiments demonstrate that NCA-based sensing networks accurately estimate object position at 0.24 times the inter sensor distance. They maintain resilience under sensor faults and noise, and scale seamlessly across varying network sizes. These findings underscore the potential of local, decentralized computations to enable scalable, fault-tolerant, and noise-resilient object property estimation in DMS
\end{abstract}


\section{Introduction}
A \ac{DMS} consists of numerous actuators, often arranged in a lattice topology, that work collaboratively to manipulate objects on its surface. Through coordinated use of many actuators, these systems can achieve precise positioning, orientation, and manipulation of objects.
This collaborative approach offers capabilities that surpass those of systems relying on independent actuators, which are often constrained by limitations in, e.g., force generation, stabilization, and actuation range \cite{bohringerDistributedManipulation2000}. Due to their spatial distribution, \ac{DMS} cover a large workspace for manipulation and the possibility of parallel manipulation \cite{reznik_cmon_2001, ataka_design_2009}.

Accurate knowledge of an object's location is crucial for orchestrating manipulation in a \ac{DMS}  using closed-loop control \cite{luntz_distributed_2001}. Estimation of an object's \ac{GC} provides a low-dimensional representation of its position, equivalent to the center of mass for homogeneous objects. This information is critical for accurate manipulation through the application of external force. Many \ac{DMS} utilize a centralized sensing system, such as single-camera computer vision systems \cite{murphey_feedback_2004}. However, centralized architectures introduce vulnerabilities: failure of a single sensor can compromise the entire system. Moreover, such sensing systems are external to the manipulator, and the need for specialized processing hardware for signal processing (e.g. FPGA), impairs their integration greatly \cite{ataka_layer-built_2007}.

Decentralization offers a promising alternative for \ac{DMS}, addressing many of the limitations of centralized systems. By distributing the computational load of sensing and control across multiple components, decentralization mitigates the scalability challenges posed by centralized control, which often suffer from computational, processing, and communication bottlenecks \cite{agarwal_velocity_1998}. Additionally, decentralization enhances robustness by eliminating the existence of a single point of failure. 
However, decentralized systems face their own challenges, particularly in coordinating interactions between agents. Inter-agent communication overhead can lead to latency and complicate real-time interactions among manipulators. %

Localized sensing can provides precise information about object-manipulator interactions at specific known points. Information of such interactions is key for accurate manipulation through controlled force application. However, such locality is absent in camera based perception system. When localized sensors are utilized in a high spatial density, the system is able to detect objects in contact with multiple sites simultaneously, allowing for calculation of  locla distributions of the sensed modality, for example pressure maps. The use of multiple sensors also enhances system robustness by preserving functionality, even in the case of individual sensor failures or noise. Despite this, existing \ac{DMS} typically implement one sensor per actuator in low density, discontinuous arrangements, limiting their ability to capture global object dynamics effectively.

Recognizing the critical need for decentralization, scalability, and robustness in \ac{DMS}, this work offers two main contributions. First, We propose a novel inductive sensor design tailored for distributed sensing, offering a scalable, fault-tolerant approach to sensing. Second, we investigate how an \acf{NCA} based system utilizing this sensor can estimate global properties from purely local information, and how such a system is scale invariant and robust to fault, both key factors for producing \ac{DMS} of any size.

\section{Related Works}

Advancing the capabilities of \ac{DMS} requires addressing challenges in sensing and control architectures. This section reviews existing approaches to sensing in \ac{DMS} and explores the potential of \ac{NCA} as a scalable and robust solution for decentralized systems.

\subsection{Sensing in Distributed Manipulation Systems}

\ac{DMS} have traditionally relied on open-loop control for object manipulation \cite{bohringer_sensorless_1995, georgilas_cellular_2015, liu_robotic_2021, liu_micromachined_1995}. Such systems often approximate programmable vector fields to achieve sensor-less object manipulation \cite{bohringer_theory_1994, bohringer_part_2000, patil_linear_2023}.

When sensing is incorporated, single-camera systems remain the most prevalent approach. Cameras provide rich environmental information, enabling object localization, pose estimation, and obstacle avoidance.
Their widespread availability, large sensing area, and relatively high information density make them an attractive choice for \ac{DMS} \cite{reznik_cmon_2001, ataka_design_2009, georgilas_cellular_2015}. Beyond standard RGB cameras, some \ac{DMS} incorporate depth cameras \cite{follmer_inform_2013}, depth cameras and RGB cameras \cite{uriarte_control_2019}, or marker-based systems \cite{xu_modular_2024} for accurate 3D tracking, particularly pertinent for systems capable of large vertical displacement. %

In contrast, other \ac{DMS} utilize localized sensing mechanisms to detect object interactions. Resistive tactile force sensors mounted on the manipulator end-effector are commonly used to measure contact forces, offering a robust and cost-effective solution \cite{robertson_compact_2019, xue_arraybot_2023}. 
Photodiodes \cite{berlin_motion_2000, ataka_layer-built_2007, bedillion_distributed_2013} and proximity sensors \cite{fukuda_hybrid_2000} have also been employed to detect position of objects.  Some systems combine local sensing with external camera-based perception to fuse modalities for enhanced object tracking \cite{parajuli_actuator_2014}. The low cost of tactile sensors makes them viable for high-density deployment, facilitating spatially distributed sensing. %
However, current \ac{DMS} designs often employ low-density, discontinuous sensor configurations, typically using a single sensor per actuator, which restricts their capacity to effectively capture the object dynamics in detail.

Effective \ac{DMS} sensing systems should provide precise local sensing with high spatial density to accurately capture global object dynamics. Scalability demands cost-effective sensors for large-scale deployment, along with architectures  capable of handling inevitable sensor failures. In this work, we propose such a system utilizing a novel inductive smart sensor surface, able to provide high precision localized sensing. We then implement a \acf{NCA} based a architecture, facilitating a purely decentralized sensing framework, that is robust to failures and capable of scaling to any scale.

\subsection{Neural Cellular Automaton}

\acf{CA} consist of a regular grid of cells, each occupying one of a limited set of states. The state of a cell is updated based on its own current state and that of its neighbors, according to predefined rules. Despite the simplicity of the local rules, \ac{CA} have been shown to give rise to very complex emergent behaviors, exemplified by Conway’s Game of Life\cite{adamatzky_game_2010}. However, crafting rules for specific behaviors is non-trivial, prompting a shift toward discovering rule sets through automated approaches \cite{wolfram2021problem}.

Recently, deep learning has been integrated into \ac{CA} to learn rule sets that drive desired behaviors through gradient-based optimization of a loss function \cite{ha_collective_2022, gilpin_cellular_2019}. Neural Cellular Automata merge \ac{CA} principles with \ac{NN}, representing the \ac{CA} update function $f_{\theta}$ as a network taking as input the agents neighborhood. A unique feature of \ac{NCA}s is the use of hidden channels in each cell’s state in which free tokens of information are stored for inter-agent communication \cite{wulff_learning_1992,mordvintsev_growing_2020}.

\acp{NCA} have been applied to a variety of tasks requiring global property estimation from local interaction. 
For instance, \cite{randazzo_self-classifying_2020} demonstrated the use of \acp{NCA} to classify MNIST digits through consensus among agents. Similarly, \cite{nadizar_fully-distributed_2023} proposes a shape-aware controller where each module infers the shape of a larger assembly. \cite{walker_physical_2022} extended this concept to modular robotics, enabling systems to infer their own shapes through local communication, and successfully transitioned from simulation to hardware.\cite{bessone_neural_2025} introduced a methodology for inferring the geometric center of objects laying on a grid of sensing agents. Leveraging the capabilities of \acp{NCA}, each agent locally shares information within its neighborhood, enabling the inference of global properties through purely local communication.
This paper builds upon these findings to address the challenges of decentralized sensing in \ac{DMS}, introducing a system that bridges the gap between simulation and real-world hardware.

\section{Soft Inductive Sensing Platform} %

To satisfy a \acp{DMS} requirements for high precision local sensing at the actuator level, we developed a soft inductive sensing layer that serves dual purposes: enabling object manipulation as the system's end effector and collecting detailed tactile data about objects interacting with the surface.

\subsection{Design Overview}

The sensor design employs inductive sensors positioned bellow a compliant surface embedded with ferromagnetic material. Deformation of the soft material under load changes the distance between sensor and ferromagnetic material, causing a measurable change in inductance proportional to deformation.
A \ac{PCB}, containing a lattice of coils connected to inductive signal conditioning chips, form an array of inductive sensors. This \ac{PCB} forms the base layer for a soft structure comprising a ferromagnetic sheet encapsulated within lightweight, compliant polyurethane foam (Poron\textregistered) and topped with a smooth, FDA-approved \ac{PET} film. This single soft structure allows these sensors to be combined into a continuous soft sensing surface. This surface is well suited as a sensorized \ac{DMS} end-effector, and for industrial applications like food handling and packaging, due to its ruggedness and resilience against dust and moisture. 
This architecture allows for low-cost scalability. The number, arrangement, shape, and spacing of coils within each \ac{PCB} configurable to application need.

\subsection{Prototype board design and manufacture}

 The sensor design utilizes a LDC1614 inductive signal conditioning chip and paired coil as an inductive sensor, following the procedure in \cite{lo_preti_sensorized_2023}. For use in in our experimental testing, a FR4 \ac{PCB} hosting 16 embedded inductive coils arranged, each measuring $30 \times 30$~mm, were arranged in a in a $4 \times 4$ grid. The tactile sensor prototype uses four LDC1614 connected to two I2C lines, with each chip polling four coils and two chips per I2C line. This configuration allows precise measurements at up to the kHz range. To optimize performance, traces between the coils and driver chips are isolated to enhance the signal-to-noise ratio, and the board is shielded from electromagnetic interference using an additional ferrite layer beneath the \ac{PCB}. In this prototype, we connect the sensors to a Teensy\textregistered micro-controller from which to read sensor data via I2C.

\begin{figure}[t]
    \centering
    \includegraphics[width=0.85\columnwidth]{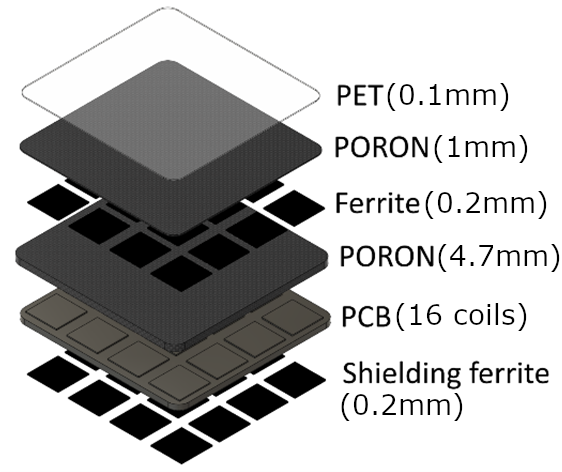}
    \caption{Exploded schematic of the soft sensor prototype, showing the PCB, ferrite, Poron\textregistered, and PET layers.}
    \label{fig:soft_sensor}
\end{figure}

The sensor's structure and material selection is illustrated in \cref{fig:soft_sensor}.
The multilayer tactile structure above the \ac{PCB} integrates two Poron\textregistered\ foam layers (1mm and 4.7mm respectively) with a ferrite sheet (0.2mm) to maximize compliance and sensitivity. 
The layers are bonded with Sil-Poxy\textregistered adhesive.
The \ac{PET} top layer (0.1mm) provides a smooth, low-friction contact surface, ideal for handling soft materials and reducing wear. Laser-cut plexi glass masks were used for precise assembly and to ensure manufacturing consistency.

\subsection{Characterization and Testing}

A three-axis indentation setup tested the tactile sensing layer prototype. Two micrometric manual linear stages controlled the positions of the X and Y stages. An M-111.1DG translation stage was positioned along the Z-axis on top of the X stage, controlled by the paired C-884.4DC motion controller (Physik Instrumente, USA). An ATI Nano17 (ATI Industrial Automation, USA) load cell was mounted on the Z stage with an L-shaped part to adjust its configuration, under which an ABS probe with a round tip shape was attached to the load cell. The prototype was placed on a lab jack (MKS Instruments, Inc.) and fixed with a base designed to locate the four testing positions. 

Key performance metrics, including repeatability, range, crosstalk, RMSE, hysteresis, and sensitivity, were analyzed for all 16 inductive coils under controlled loading conditions. Repeatability in all coils was measured at 68.22~\% with a standard deviation of 27.33~\%. The maximum force range was 6.72~N $\pm$ 0.78~N, and crosstalk between neighboring coils was minimal, measured at 1.57~\% $\pm$ 1.37~\%.

To generate calibration curves, we applied a uniform pressure to the (30 \(mm^2\)) area above each sensor via incremental loading, applying between 0.04905 - 5.886~N of force, equivalent to 5-600~g of applied mass. 
Polynomial curve fitting per coil found each coil's response to be predominately linear, though distinct due to edge effects effects and inbuilt manufacturing variability.

\section{Decentralized Sensing} %

\begin{figure*}[th]
\begin{subfigure}{0.5\textwidth}
\centering
\includegraphics[height=6cm]{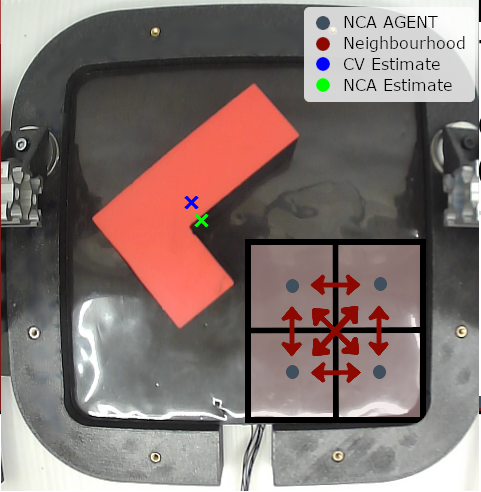}
\captionsetup{width=0.9\textwidth}
\caption{Geometric center of object detected by computer vision system (blue) and calibrated NCA estimate, projected to object top surface (green). Visualization of neighborhood (red) of agents (gray).}
\label{fig:estimate_projections}
\end{subfigure}
\begin{subfigure}{0.5\textwidth}
\centering
\includegraphics[height=6cm, trim={5 5 5 5},clip]{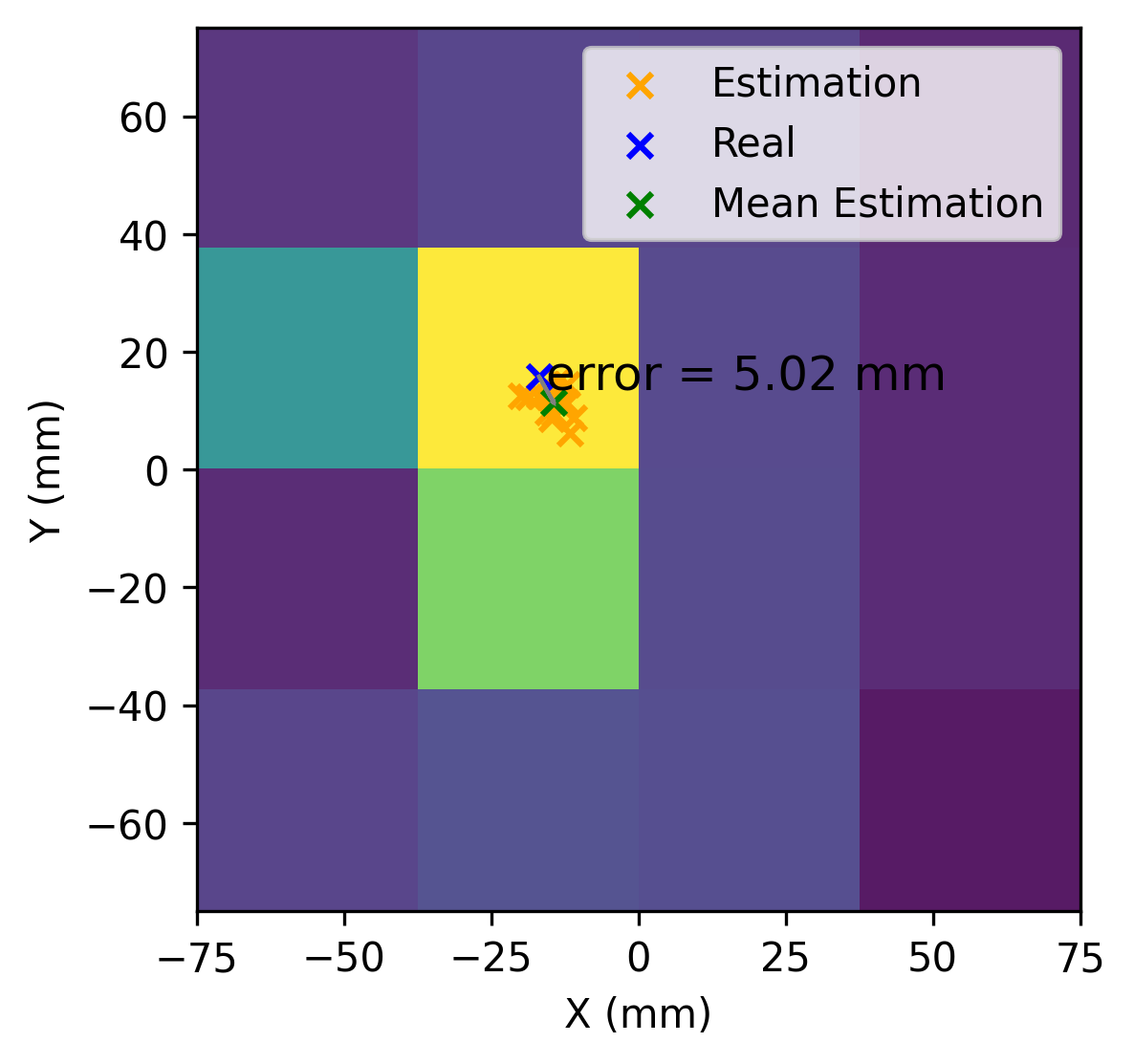}
\captionsetup{width=0.9\textwidth}
\caption{Estimates for each of the NCA agents (orange), their mean (green), and object center detected by computer vision system (blue)\\
}
\label{fig:NCA_estimates}
\end{subfigure}

\caption{Estimation of Geometric Center for object in contact with sensing surface }
\label{fig:twin_image}
\end{figure*}

In this work, we estimate the global properties of objects in contact with the tactile sensing layer, specifically the \ac{GC} of the surface of an object in contact with the sensor. For objects of uniform density, this corresponds to the 2D projection of their \ac{CoM}.

\subsection{Data collection}

To train the \ac{NCA} model, a dataset was created containing two key components: readings from the sensors when objects were in contact with the sensing surface, serving as input, and the ground truth geometric center of each object, serving as the target output. The ground truth was determined using a computer vision system developed for this purpose using the OpenCV library\cite{bradski_opencv_2000}.

\begin{figure*}[t]
\centerline{\includegraphics[width=0.85\linewidth]{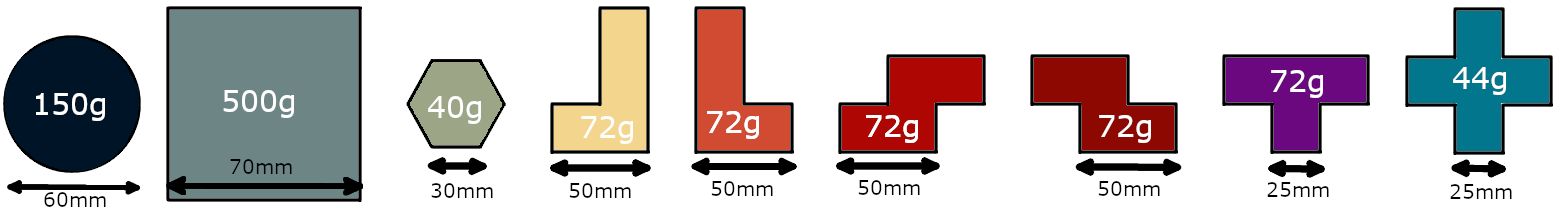}}
\caption{Contact footprint and mass of objects used to create dataset.}
\label{fig:shapes}
\end{figure*}

The dataset consisted of distinct geometric objects with uniform mass distribution but varying shape and mass, as detailed in \cref{fig:shapes}. During data collection, a predetermined face of each object was placed in direct contact with the sensor surface. Sensors readings were sampled at 20~Hz for 2.5 seconds to produce 50 samples per object per position. Between 50-150 positions were recorded for each object, depending on object size relative to sensor area,  across the entire sensor surface.

To ensure reliable detection by the  computer vision system, all objects were 3D printed from bright, mono-colored PLA. Edges of the objects not in contact with the sensor or relevant for detection were masked with black tape. Images were captured under controlled lighting conditions, and the OpenCV Python library was employed to calculate the geometric center of the face of the object in contact with the sensor. The geometric center was then mapped to the coordinate frame of the sensor board, providing ground truth data for training. 

\subsection{Neural Network Model}
\label{sectionNNmodel}
A decentralized system was implemented, where an array of \ac{NCA} agents received inputs directly from individual sensors. The spatial distribution of the sensors, such as those within the sensor board mirrors the lattice structure of a 2D \ac{CA}, enabling decentralized and spatially distributed sensing.

Although the sensors were implemented within the same sensor board for manufacturing simplicity, each \ac{NCA} agent could only access the local information of its corresponding sensor and its neighbors. This collectively forms a distributed network where the agents rely solely on local information; a fully decentralized computational paradigm. 

In the \acp{NCA} framework, each agent maintains a state $S$ that evolves iteratively through an asynchronous update process governed by a \ac{NN}-based update function. The state $S$ of a tile $i$ at time $t$ is updated according to:

\begin{equation}
    S_{i}^{t+1} = f_{\theta} ( \{ S_{j}^{t} \}_{j\in N(i) } )
\end{equation}

where $f_\theta$ is the additive update function parameterized by $\theta$, this function takes as input the states of tile $i$ and its neighborhood $N(i)$ from the previous time step $t-1$, enabling localized updates influenced by both the tile itself and its neighbors.

\subsubsection{Agents State}
The state $S$ of each \ac{NCA} agent encapsulates multiple components: the sensor value $V$, which captures tactile interactions with the environment; the global property estimation $E$, representing the agent's prediction of the global property; a set of hidden channels $H$, which serve as auxiliary memory or communication channels; and information from its neighborhood $N$, which encodes the states of the tiles within the agent's Moore's neighborhood, as illustrated in Fig. \ref{fig:twin_image}. 
The neighborhood $N$ is restricted to immediate neighbors and does not extend to the neighbors of neighbors. 

During training, the \ac{NN} modifies $E$ to minimize the prediction error, while the dynamics of $H$ are left to emerge as the network optimizes its functionality. In this framework, global consensus emerges iteratively through local exchanges, with agents requiring multiple update steps to converge. The number of iterations is randomly sampled from a uniform distribution in the arbitrary range of 15-30 time steps to ensure robustness to long-term stability issues, as described in \cite{wolfram_universality_nodate}.

In a distributed setting, a shared global clock cannot be assumed, in such scenarios, the agents update their states asynchronously. During each training step, only a randomly selected subset of agents updates its state. Once the predetermined number of iterations is reached, the estimation error is calculated as the mean Euclidean distance between the predicted center of the object $(x_{Ei}, y_{Ei})$ for each agent and the actual center $(x_C, y_C)$ determined via a computer vision model. Implementation and reproducibility kit available in the footnote \footnote{https://github.com/nhbess/NCA-REAL}. %

\subsubsection{Architecture}

The update function $f_\theta$ is implemented as a \ac{NN} with three main layers: The Perception Layer, which applies a $3 \times 3$ convolutional kernel to extract local features, and a Sobel filter to compute gradients of the states along the $x$ and $y$ axes; a Processing Layer utilizing a $1 \times 1$ kernel to reduce dimensionality and extract relevant features, with a with a Rectified Linear Unit (ReLU) to introduce linearity; finally, the Output Layer, also employing a $1 \times 1$ kernel, generates residual updates to the agent's state, modifying only the global property estimation and hidden channels while preserving other components of the state.

\subsubsection{Training Methodology}

Training the \ac{NCA} involves learning the parameters $\theta$ of the update function $f_\theta$ to ensure that the estimated global property $E$ converges to its true value. All agents in the system are identical and share the same neural network. The was randomly divided into two equally sized distinct sets: training and testing. The key variable of interest, the estimation $E$, is used to compute the loss function by comparing agent estimation to the true center of the object derived from the dataset.

To enhance stability, the training incorporates a pool-based strategy, in which poorly performing states are periodically replaced with empty states from the pool, as detailed in \cite{mordvintsev_growing_2020}. %
This approach mitigates training instability and ensures robust performance. The efficacy of this methodology has been demonstrated previously \cite{bessone_neural_2025}, validating its application in distributed sensing.

\section{Experiments}

To assess the performance of our system, in context of a \ac{DMS} as previously described, we conducted a series of experiments. First, we aim to quantify the accuracy of the decentralized estimation of the global property: the geometric center. Additionally, we investigate the system's robustness, which we categorize into two distinct aspects: fault tolerance and noise tolerance. Fault tolerance pertains to the system's ability to maintain functionality when individual components fail or are removed, while noise tolerance evaluates its resilience to variations or noise in sensor signals. Lastly, we explored the scalability capabilities of the model by testing its performance across sensor arrays of varying sizes. %

\subsection{Performance Evaluation}
\label{ss:perfeval}

\begin{figure}[t]
\centerline{\includegraphics[width=0.85\columnwidth]{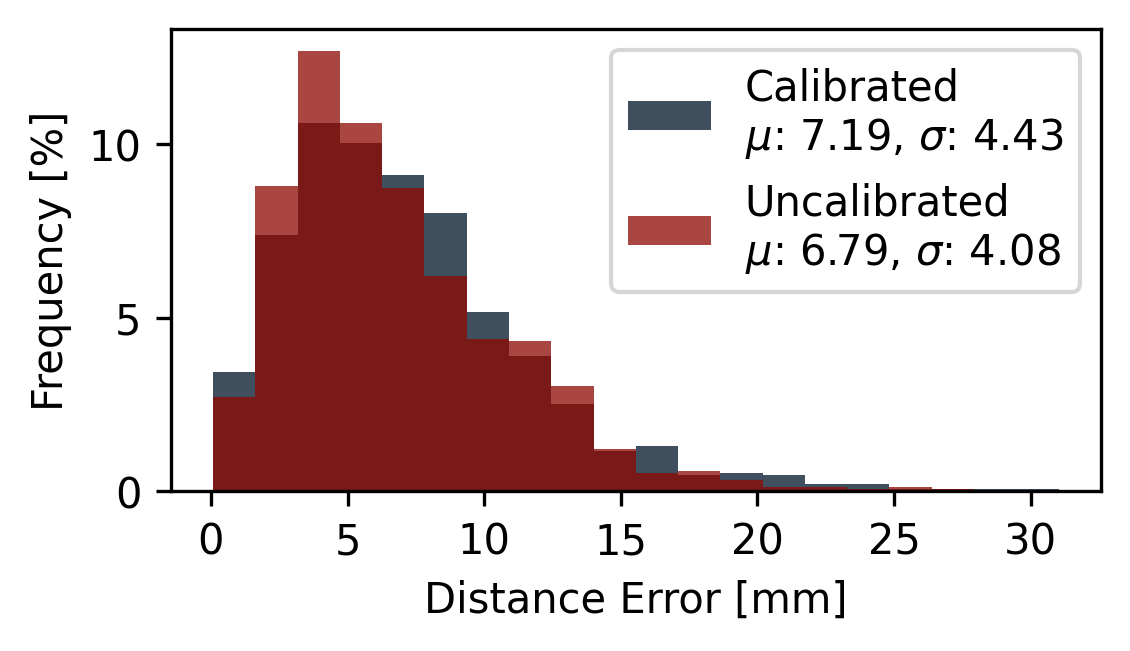}}
\caption{Error distribution of models trained and evaluated with calibrated and uncalibrated data from respective test sets.}
\label{fig:comparison}
\end{figure}

In this experiment, we compared the performance of two models, trained using calibrated and uncalibrated data, respectively, as detailed in \cref{sectionNNmodel}. Both models were tested on their respective datasets, and the average estimation provided of individual agents was computed with the ground truth. \Cref{fig:comparison} shows that both models exhibit an average distance error of $7.19$ mm and $6.79$ mm respectively.

This result highlights that the system can effectively estimate the \ac{GC} with a high degree of accuracy. Given the coil width of 30~mm and a center-to-center coil spacing of 37.5~mm, achieving a positional error of below 7.2 mm ($\approx$ 0.24 times the width of a coil) showcases the system’s ability to accurately determine the \ac{GC} at much smaller distances than coil spacing of without relying on it being positioned directly above a single coil. 

The use of uncalibrated data appeared to slightly reduce estimation errors compared to the use of calibrated data. However, after performing statistical testing, the distributions were found not to be statistically significant, with $P= 0.304$.

\subsection{Comparison with Centralized Approaches}

\begin{figure}[t]
\centerline{\includegraphics[width=0.85\columnwidth]{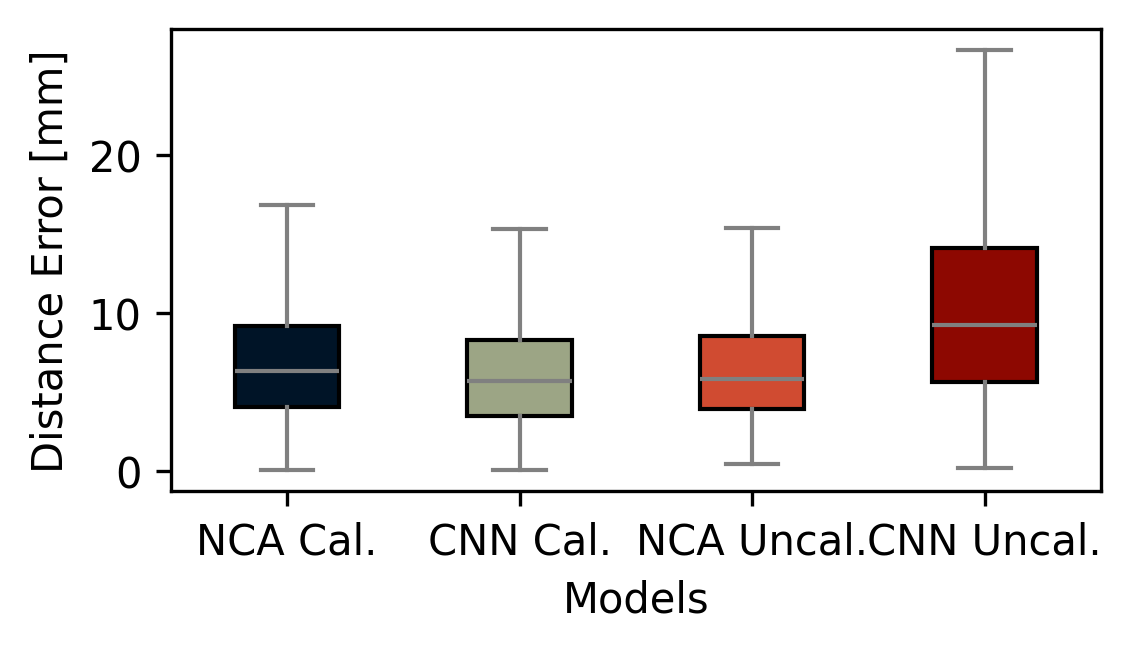}}
\caption{Box plots of estimation error for Centralized and NCA approaches, showing median (line), inter-quartile range (box), and range (whiskers)}
\label{fig:center}
\end{figure}

To evaluate the performance of the proposed \ac{NCA}-based system, we compare it against a centralized \ac{CNN} architecture. The centralized \ac{NN} processes all sensor readings simultaneously using two consecutive convolutional layers to extract global spatial features, followed by 3 fully connected layers to estimate the object's position. The resulting comparison is shown in \cref{fig:center}, where it can be observed that both the centralized and decentralized approaches perform comparably on calibrated data. Notably, the uncalibrated centralized model exhibits slightly worse performance, potentially due to domain shifts that the architecture cannot accommodate without calibration.

\subsection{Fault Tolerance}

\begin{figure}[t]
\centerline{\includegraphics[width=0.85\columnwidth]{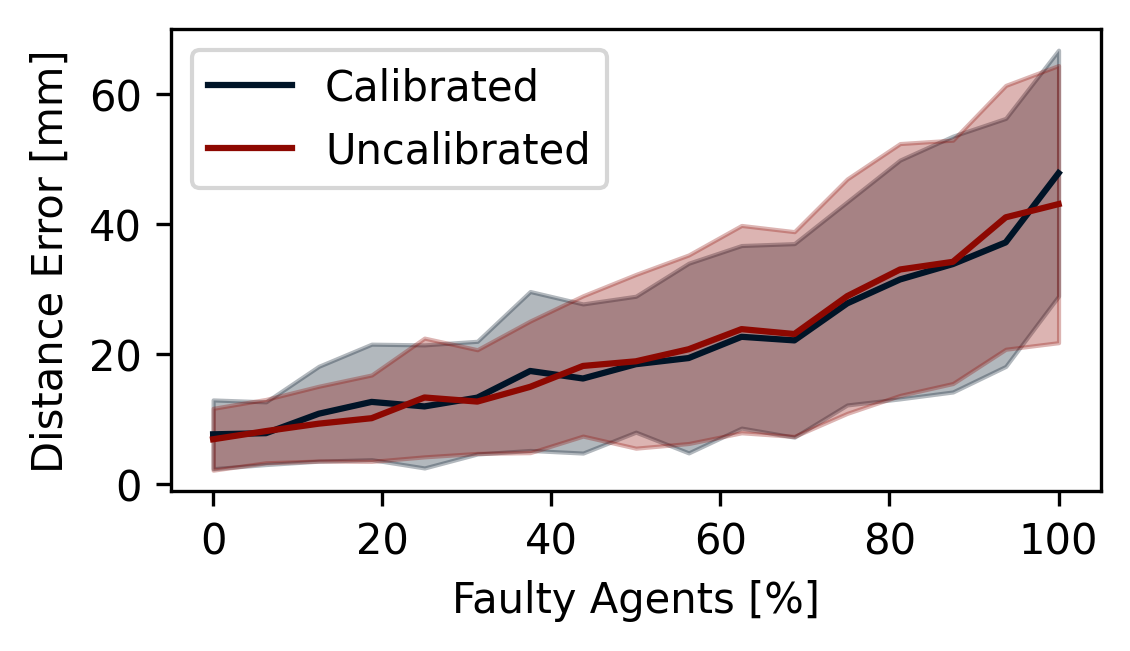}}
\caption{Mean and standard deviation of estimation error for different percentages of faulty agents.}
\label{fig:faulty}
\end{figure}

Fault tolerance was evaluated by systematically introducing sensor failures and measuring the system's performance.  Faulty sensors were simulated by replacing their readings with zero, effectively removing their influence on the system's computations. We masked a fault on between $0\%$ to $90\%$ of sensors (rounded down) in increments of $10\%$. 
    The model was then evaluated across $100$ different object locations, each time with a new random selection of faulty sensors to ensure statistical robustness. We measured the distance error between the estimated \ac{GC} and the ground truth, recording both the mean and standard deviation of these errors, the results are shown in \cref{fig:faulty}.

\subsection{Noise Tolerance}

The system's noise tolerance was evaluated by introducing Gaussian noise to sensor data at varying levels, ranging from $0\%$ to $100\%$ in increments of $10\%$. The noise magnitude was scaled relative to the sensor's signal strength, simulating real-world conditions such as electronic interference, signal attenuation, or sensor drift.

\begin{figure}[t]
\centerline{\includegraphics[width=0.85\columnwidth]{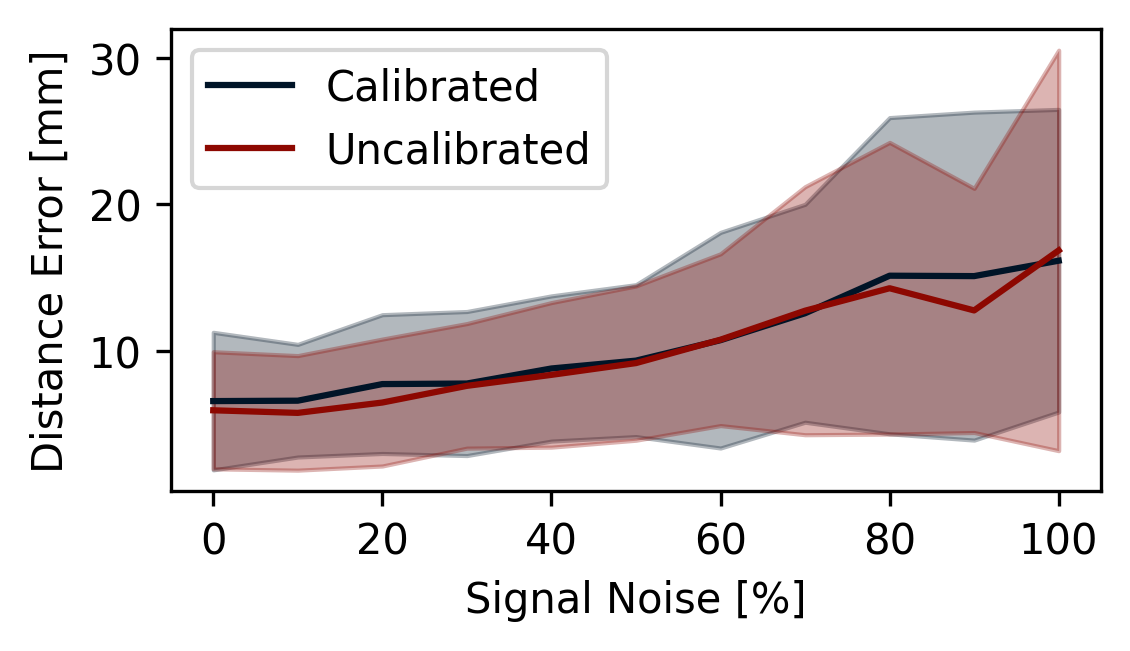}}
\caption{Mean and standard deviation of estimation error for different noise levels.}
\label{fig:noise}
\end{figure}

For each noise level, 100 trials were conducted, and the mean and standard deviation of the distance error were recorded. As shown in \cref{fig:noise}, the system demonstrates resilience to increasing noise levels, maintaining consistent accuracy at moderate noise intensities and gracefully degrading as noise becomes severe. 

\subsection{Scalability}

One of the key advantages of \acp{NCA} is their inherent locality, allowing the same trained model to be applied to networks of varying sizes without retraining. 

\begin{figure}[t]
\centerline{\includegraphics[width=0.85\columnwidth]{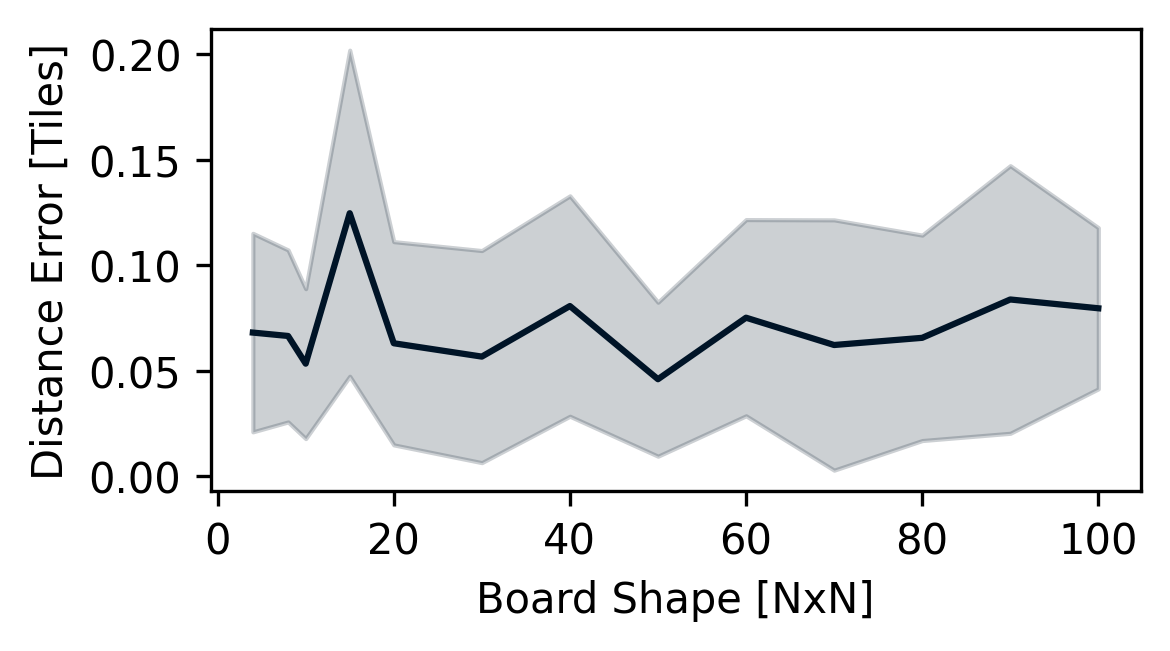}}
\caption{Mean and standard deviation of estimation error for different sizes of agents network.}
\label{fig:scale}
\end{figure}

To evaluate this property, a model trained on a grid of $8\times8$ sensors was tested on grids ranging from $4 \times 4$ to $100 \times 100$, representing up to 10000 agents. 
Due to hardware limitations, binary synthetic data was used for both training and evaluation. 
Performance was measured in terms of distance errors expressed in tile sizes (the distance between adjacent sensors). \Cref{fig:scale} shows the results of the experiment.

The mean distance errors remained consistent across all network sizes. Error metrics did not increase with the addition of more agents, indicating that the model successfully scaled without any degradation in performance. This consistency highlights the effectiveness of \acp{NCA} in maintaining robustness and accuracy, regardless of the network's size.

\section{Discussion}

\subsection{Centralized vs Decentralized sensing}

There are fundamental trade-offs between centralized and decentralized architectures for global property estimation. Our decentralized \ac{NCA}-based approach performs comparatively to a centralized baseline approach. Centralized systems necessitate all sensor data is aggregated, creating a computational bottleneck and a single point of failure, which our system avoids. However, this does allow rapid access to information from across the system, without the need to wait for information propitiation via local communication, which is costly both in time and computational overhead. This delay can be seen in the non-instantaneous nature of generating our system's estimate. Therefore, our system may struggle to achieve optimal results in the case of highly dynamic environments where object properties of position varied rapidly. 

Decentralized perception systems also face complexity challenges, as they lack direct access to global information and cannot implement straightforward global algorithms for property estimation. While decentralized variants of such algorithms have been explored \cite{bedillion_distributed_2013}, they incur additional computational costs. 
Despite these challenges, the scalability and robustness inherent to decentralized architectures make our \ac{NCA}-based approach a promising candidate for large-scale \ac{DMS}, particularly in harsh environments with high noise levels or a high likelihood of sensor failure. 

\subsection{Sensor Calibration}

 The similarity of models trained on both calibrated and uncalibrated sensor data suggests that, in this context, the calibration process may not substantially influence the system’s estimation capabilities. If calibration is unnecessary, this reduces the complexity of setup for any practical deployments. However, although we have shown this state to be the case in this context, this state does not necessarily hold in general. Sensors manufactured with a higher manufacturing variability (i.e. not manufactured as part of the same sensor board) or operating in substantially different conditions, which are not accounted for without calibration. Soft sensors may be especially susceptible to these variations,  due to the large mechanical deformations they experience in use. %

Results from the noise tolerance experiments provide insights into valid operational conditions. By examining how the system responds to controlled variations in sensor input quality, it is possible to approximate the influence of different calibration distributions on estimation error, thus providing a measure of confidence in the model’s robustness when used with diverse sensor configurations.

\subsection{System Robustness}

The system’s fault tolerance experiment reveals that meaningful performance is maintained even when up to $30\%$ of the sensors are rendered non-functional, and a robustness up to a $50\%$ signal-to-noise ratio indicates a reliability to the system. In any real world implementation, such noise and sensor failures are inevitable, especially as the number of sensors increases. Such deviation in sensor readings would also be expected on a non-static system, as in a \ac{DMS} actuator, in which measured pressure applied to the sensor board would vary as the system moves. Therefore, this consistency in estimation offers validation for the utility of such a system for real world application.
The removal of a centralized processing unit significantly enhances robustness by eliminating a single point of failure. The consensus mechanism inherent to \ac{NCA}s ensures that the system remains robust even when individual agents receive corrupted information. This resilience to sensor failures and noisy inputs is critical for real-world applications, where unpredictable environmental factors and hardware degradations are common.

\subsection{Scalability}
One of the most notable advantages of the proposed \ac{NCA}-based approach is its scalability. The system's performance remains consistent across varying network sizes by relying solely on local information and decentralized decision-making. Unlike centralized architectures, this system imposes no size limitations due to computational overhead, allowing for theoretically unlimited scaling. 
However, as the number of agents increases, so does the time for information to propagate via local communication. This limitation could potentially be exploited, prioritizing sensing and manipulation for objects in the immediate vicinity of an agent in the case of mutli-object manipulation.
Additionally, the ability to implement a single agent on multiple system scales without need for retraining offers a massive benefit for practical deployment. This locality also indicates such a architecture would be well suited for modular designs, where the sensing surface can be easily adapted with little need for model retraining.

\section{Future Work}

\subsection{Non-static objects and surfaces}
While the experiments in this study focused on static objects, real-world applications in object manipulation involve dynamic environments. A static approximation may hold with a sufficiently high sampling rate relative to object speed.
However, challenges arise from the time required for information propagation and consensus formation in decentralized systems.
When integrated as a robotic end effector, the sensor will be non-static and will experience induced force, altering estimation. Even if held static, the surface will experience pose variations that influence contact forces. While traditional mechanics can estimate these forces, additional complexities, such as soft-surface compression under non-perpendicular orientations to gravity, must be considered. Although noise robustness testing offers some insight into the behavior behavior of the current system, addressing these effects will require further experimentation and enhanced model training.

\subsection{Tunable Material}
The materials of the soft surface influence both sensor characteristics and manipulation capabilities. Tailoring these properties for specific applications, such as optimizing deformation to maximize the dynamic sensing range for objects with known properties, holds significant potential. Furthermore, varying material properties across the surface could enable new sensing and manipulation strategies, which we intend to explore in future work.

\section{Conclusion}

This work presents a decentralized strategy for sensing in \acf{DMS} using \acf{NCA} in combination with a newly developed soft inductive sensor. We demonstrated that this approach reliably infers global object properties, such as the geometric center, using purely local interactions. The system exhibits robustness to noise and sensor faults, maintaining performance under challenging conditions. 
Moreover, we have shown that these models scale without loss of accuracy to significantly larger sensor arrays, reinforcing the potential of decentralized methodologies to overcome the limitations of centralized sensing and control. This proof of concept lays the foundation for fully distributed, fault-tolerant manipulation architectures apable of dynamically responding to component failures.
The findings suggest a promising direction for future research in developing versatile, scalable, and efficient \ac{DMS}, particularly for applications in harsh environments or those requiring high reliability and adaptability.

\subsection{Acknowledgments}
This work was conducted as part of the MOZART project, funded by the European Union, EU project id: 101069536.

\bibliography{references}
\bibliographystyle{IEEEtran}
\end{document}